\documentclass[runningheads]{llncs}
\usepackage{amssymb}
\usepackage{amsmath}
\usepackage{graphicx}
\usepackage{subcaption}
\usepackage{hyperref} 
\usepackage{float}
\usepackage{calrsfs}
\usepackage{graphicx}
\usepackage{tikz}
\usepackage{wrapfig}
\usepackage[T1]{fontenc} 

\usepackage{array}
\usepackage{pifont}
\DeclareMathAlphabet{\pazocal}{OMS}{zplm}{m}{n}

\usepackage{multirow}
\usepackage{caption}
\newcommand{\InsetArchDiagram}{
\begin{figure}[h]
\begin{tikzpicture}
    \node[anchor=south west,inner sep=0] at (0,0) { %
        \includegraphics[width=\linewidth,page=1,trim={0cm 0cm 0cm 0cm},clip]{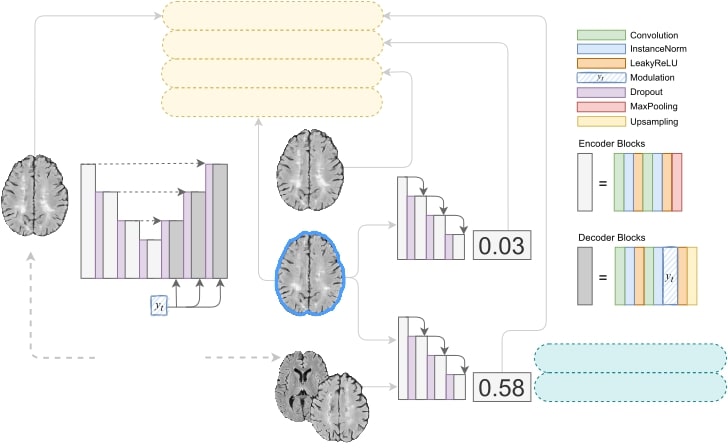}
        }; 
    \node[anchor=south west] at (2.3, 1.4) {$\pazocal{X}$};

    \node[scale=0.8, anchor=south west] at (1.7, 5) {\tiny Generator $G(\cdot, \cdot)$};
    \node[scale=0.65, anchor=south west] at (1.05, 4.8) {\tiny $(\mathbf{R}^{w_0 \times h_0 \times d_0} \times \{0, 1\}) \rightarrow \mathbf{R}^{w_0 \times h_0 \times d_0}$};
    \node[anchor=south west] at (0.35, 3.25) {\tiny $\mathbf{x}$};

    \node[anchor=south west] at (4.7, 3.65) {\tiny $G(\mathbf{x}, 1)$};
    \node[scale=0.7, anchor=south west] at (4.3, 3.6) {\tiny Active reconstruction};
    \node[anchor=south west] at (4.7, 1.9) {\tiny $G(\mathbf{x}, 0)$};
    \node[scale=0.7, anchor=south west] at (4.2, 1.85) {\tiny Inactive counterfactual};
    \node[anchor=south west] at (4.5, 0.1) {\tiny $\mathbf{x}_{y=0}$};
    
    \node[anchor=south west] at (0.6,1.1) {\textcolor{gray}{\tiny $\mathbf{x} \sim \pazocal{X}_{y=1}$}};
    \node[scale=0.7, anchor=south west] at (0.6,1.5) {\textcolor{gray}{\tiny Real active samples}};
    \node[anchor=south west] at (3.2,1.1) {\textcolor{gray}{\tiny $\mathbf{x} \sim \pazocal{X}_{y=0}$}};
    \node[scale=0.7, anchor=south west] at (3,1.5) {\textcolor{gray}{\tiny Real inactive samples}};

    \node[scale=0.8, anchor=south west] at (6.9, 0.5) {\tiny Discriminator};
    \node[scale=0.65, anchor=south west] at (6.6, 0.3) {\tiny $D( \cdot): \mathbf{R}^{w_0 \times h_0 \times d_0} \rightarrow [0, 1]$};

    \node[scale=0.8, anchor=south west] at (7.05, 2.9) {\tiny Classifier};
    \node[scale=0.65, anchor=south west] at (6.8, 2.7) {\tiny $f( \cdot): \mathbf{R}^{w_0 \times h_0 \times d_0} \rightarrow [0, 1]$};

    \node[scale=0.9, anchor=south west] at (3,7) {\tiny $\pazocal{L}_{d}(G, D) = \mathrm{BCE}[D(G(\mathbf{x},0)), 1] $};
    \node[scale=0.9, anchor=south west] at (3,6.55) {\tiny $\pazocal{L}_{c}(G, f) = \mathrm{BCE}[f(G(\mathbf{x},y_t)), y_t] $};
    \node[scale=0.9, anchor=south west] at (3,6.05) {\tiny $\pazocal{L}_{vox+}(G) = \mathrm{MSE}[\mathbf{x} ,  G(\mathbf{x}, 1)]$};
    \node[scale=0.9, anchor=south west] at (3,5.6) {\tiny $\pazocal{L}_{vox-}(G) = \mathrm{MSE}[\mathbf{x} ,  G(\mathbf{x}, 0)]$};
    \node[anchor=south west] at (3.5,7.5) {\tiny \textcolor{brown}{\textit{Generator losses}}};

    \node[scale=0.85, anchor=south west] at (9,1.3) {\tiny $\pazocal{L}_{d-}(G, D) = \mathrm{BCE}[D(G(\mathbf{x},0)), 0] $};
    \node[scale=0.9, anchor=south west] at (9,0.8) {\tiny $\pazocal{L}_{d+}(D) = \mathrm{BCE}[D(\mathbf{x}_{y=0}), 1] $};
    \node[anchor=south west] at (9.4,0.45) {\tiny \textcolor{cyan}{\textit{Discriminator losses}}};
\end{tikzpicture}
\caption{Network Architecture}
\label{fig:NetworkArchitecture}
\end{figure}
}

\begin{document}
\title{Counterfactual Image Synthesis for Discovery of Personalized Predictive Image Markers}

\titlerunning{Counterfactual for discovering image markers}
%
\author{Amar Kumar\inst{1,2}
 \and
Anjun Hu \inst{1,2}
\and
Brennan Nichyporuk\inst{1,2}
\and
Jean-Pierre R. Falet \inst{1,2,3}
\and
Douglas L. Arnold \inst{3,4}
\and
Sotirios Tsaftaris \inst{5,6}
\and
Tal Arbel\inst{1,2}
}
\authorrunning{A. Kumar et al.}
%
%
\institute{Center for Intelligent Machines, McGill University, Canada
 \and
MILA (Quebec Artificial Intelligence Institute), Montreal, Canada
\and
Department of Neurology and Neurosurgery, McGill University, Canada
\and
NeuroRx Research, Montreal, Canada
\and
Institute for Digital Communications, School of Engineering, University of Edinburgh, UK
\and
The Alan Turing Institute, UK\\
\email{amarkr@cim.mcgill.ca}
}
\maketitle 

\begin{abstract}
The discovery of patient-specific imaging markers that are predictive of future disease outcomes can help us better understand individual-level heterogeneity of disease evolution. In fact, deep learning models that can provide data-driven personalized markers are much more likely to be adopted in medical practice. In this work, we demonstrate that data-driven biomarker discovery can be achieved through a counterfactual synthesis process. We show how a deep conditional generative model can be used to perturb local imaging features in baseline images that are pertinent to subject-specific future disease evolution and result in a counterfactual image that is expected to have a different future outcome. Candidate biomarkers, therefore, result from examining the set of features that are perturbed in this process. Through several experiments on a large-scale, multi-scanner, multi-center multiple sclerosis (MS) clinical trial magnetic resonance imaging (MRI) dataset of relapsing-remitting (RRMS) patients, we demonstrate that our model produces counterfactuals with changes in imaging features that reflect established clinical markers predictive of future MRI lesional activity at the population level. Additional qualitative results illustrate that our model has the potential to discover novel and subject-specific predictive markers of future activity.

\end{abstract}
\section{Introduction}
\begin{figure}[h]
\begin{subfigure}[b]{0.5\textwidth}
\centering
\scriptsize
\includegraphics[width=0.7\textwidth]{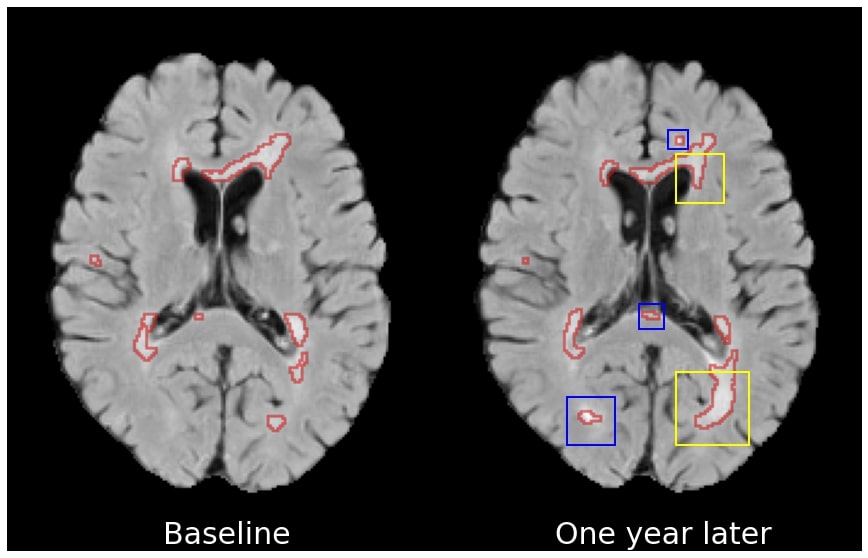}
\caption{}
\label{fig:ablations}
\end{subfigure}
\hfill
\begin{subfigure}[b]{0.5\textwidth}
\centering
\scriptsize
\includegraphics[width=0.7\textwidth]{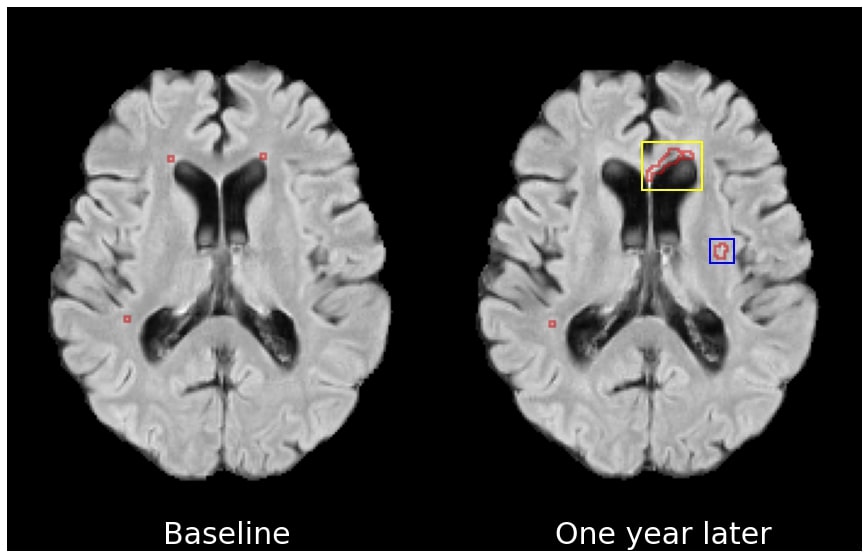}
\caption{}
\label{fig:future}
\end{subfigure}
\caption{Examples of \textit{active} patients with (a) high and (b) low lesion loads, shown at baseline and a year later. T2 lesions outlines are shown in at each timepoint (red), and new (blue boxes) and enlarging (yellow boxes) lesions are shown in the subsequent timepoint. The patient in (a) has high lesion load at baseline, and develops new (de novo) and enlarging lesions a year later, aligned with clinical expectations. The patient in (b) has low lesion load at baseline yet also develops new (de novo) and enlarging lesions a year later, defying groupwise expectations. Here, discovery of subject-specific predictive markers would be informative, and would enable image-based personalized medicine.}
\label{fig:activityLesionLoad}
\end{figure}

In many clinical contexts, treatment decisions are based on image-derived markers that are predictive of future disease activity/severity at the population level. In the context of MS, for example, it is well-established that, generally, larger T2 lesion volumes at baseline are associated with future MRI lesional activity, defined as the appearance of new (de novo) or enlarging T2 lesions (NE-T2) or Gadolinium-enhancing (Gad) lesions (surrogates of clinical relapses)~\cite{hartung2015predictors}. However, these markers do not capture the heterogeneity that exists at the individual level (e.g. patients with low lesion load can also develop NE-T2 lesions, Figure~\ref{fig:activityLesionLoad}), thus limiting the quality of personalized clinical decision-making. Therefore, identifying imaging markers predictive of an individual patient's future disease outcome would be of great importance, permitting a better understanding of personalized disease evolution in a heterogeneous population, paving the way for image-based personalized medicine. 

Recent work on explainable deep learning models have provided visual explanations for deep learning decisions, for example, to expose regions or features attended to by the network when making a decision (e.g.\ GRAD-CAM~\cite{selvaraju2017grad,jiang2020multi,panwar2020deep}, feature attribution~\cite{Baumgartner2018visualfeatureattribution,goyal2019counterfactual}). Other work disentangles clinically significant attributes from imaging modalities~\cite{xia2019pseudohealthybrain}, exploring causal inductive biases in counterfactual image generation~\cite{SCMforMS}, or generating adversarial examples~\cite{charachon2021visual,siddiquee2019learning,zia2022vant,Gifsplanation} to provide visual explanations from the differences between the input (factual image) and its counterfactual counterpart. Nevertheless, previous work has focused on examining changes in known imaging markers, for example, monitoring changes in the hippocampus for Alzheimer's patients, or producing synthesized patient images by inducing changes in lesion loads~\cite{charachon2021visual,AD,SCMforMS,Baumgartner2018visualfeatureattribution,pawlowski2020deep}, rather than on the discovery of novel patient-specific imaging markers. In fact, validation of discovered markers is usually performed based on detection overlap with known imaging markers. Furthermore, explainable models are generally focused on explaining decisions made at the current timepoint. Discovery of predictive markers for future patient outcomes remains an open problem.

In this work, we seek to discover new personalized predictive markers of disease evolution by asking the specific question: {\it What changes in the patient's current (baseline) image would lead to a different future disease outcome?} To this end, we present an {\it explainable} generative deep learning model that enables the identification of candidate biomarkers through a counterfactual synthesis process. This process involves perturbing local imaging features in baseline images that are predictive of future disease evolution. Evaluations are performed on a proprietary, large-scale, multi-center, multi-scanner, clinical trial MRI dataset acquired from patients with RRMS. Resulting counterfactuals (1) maintain subject identity by showing high structural similarity between real images and counterfactual images, (2) are realistic in terms of adherence to clinically established population-level markers (in this case, counterfactuals in part reduce baseline T2 and Gad lesion load), without \textit{a priori} awareness of these markers, and (3) have the potential to enable discovery of novel, subject-specific, local imaging markers predictive of future lesional activity, which can be visualized through qualitative analysis of the difference maps between the real baseline images and the synthesized counterfactual images.

\section{Methods}
We propose an explainable model for predicting future patient disease outcomes applied to multiple sclerosis. Explainability is achieved through a conditional generative model~\cite{van2021conditional} that synthesizes a counterfactual baseline MRI, one with an opposing future disease outcome. 

\label{sec:networkarch}
The proposed network and optimization objectives, shown in Figure~\ref{fig:NetworkArchitecture}, are designed to encourage the following characteristics of the counterfactual: (1) faithfulness to the subject (i.e.\ minimal structural perturbations to the input image), (2) adherence to the target (counterfactual) class, and (3) realism in appearance. 
\InsetArchDiagram
The network consists of three modules: (1) a generative module $G(\cdot, \cdot)$ that perturbs factual baseline MRI volume $\mathbf{x}$ to transition its class membership to a target output class $y_{t}$; 
(2) a binary future lesion activity classifier $f(\cdot)$ to ensure that the counterfactual output scans have the expected target class membership and 
(3) a discriminator $D(\cdot)$ to promote realism of the synthesized images through an adversarial training procedure~\cite{goodfellow2014generative}. 
During training, the network learns from image-space observations (baseline MRI scans) $x \in \mathrm{R}^{w_0 \times h_0 \times d_0} $ with binary class membership labels that corresponds to future lesion activity status  $y \in  \{0, 1\}$ sampled from a distribution $(x, y) \sim (\pazocal{X} \times \pazocal{Y})$.  In the context of MS, a positive class membership refers to an \textit{active} future lesion status, that is, a subject that has at least one of the following at a future time point (e.g. one year later): (1) Gad lesion(s) on T1w MRI; (2) new (de novo) or enlarging T2-weighted lesion.
In the context of biomarker discovery, the input sample to the generator is restricted to baseline MRIs of subjects that belong to the \textit{active} class with $y = 1$. This is done as validating the counterfactual of the \textit{active} class is comparatively easier than validating the counterfactual of the \textit{inactive} class. Besides, generating a counterfactual for the \textit{inactive} class is considered a many-to-one problem (i.e, multiple possible counterfactuals fit as a possible solution). This also makes the problem ill-posed and isn't actionable in a clinical context as it would imply discovering markers of future inactivity.

The conditioning mechanism permits the generative model to operate in two modes: (1) In the case that the target output class is inactive ($y_{t} = 0$), the model performs counterfactual synthesis. The model output $G(\mathbf{x}, 0)$ should be a minimally perturbed version of the input scan, with removal or modification of the imaging markers that were predictive of the future activity of this patient. This operation transitions the subject to the \textit{inactive} class while preserving the subject's identity. (2) In the case that the target output class is active ($y_{t} = 1$), the model performs factual reconstruction. The output $G(\mathbf{x}, 1)$ should replicate the input scan to the maximum extent possible while maintaining the \textit{active} class membership. 
The subject faithfulness requirement is maintained by a pair of voxel space MSE losses $\pazocal{L}_{vox+}(G)$ and $\pazocal{L}_{vox-}(G)$.  Counterfactual validity in terms of class membership ~\cite{mothilal2020explaining} is enforced through a binary cross entropy loss $\pazocal{L}_{c}(G, f)$ on a pretrained classifier $f$ (which remains frozen during the training process of the generator). 

The discriminator ensures that the  synthesized counterfactuals are also realistic, in that they should be indistinguishable from real scans sampled from the real inactive distribution. Distributional proximity between factually active scans and their $y_t=1$ reconstructions is naturally ensured by $\pazocal{L}_{vox+}(G)$. 
The synthesized counterfactuals are required to be realistic, a requirement that is enforced by $ \pazocal{L}_{d}(G, D)$ which comes from jointly training a discriminator on the inactive subset of the dataset.

The overall loss functions can be summarized as follows:
\begin{align}\label{eq:objective}
    \pazocal{L}_G = \lambda_{vox+}\pazocal{L}_{vox+}(G) + \lambda_{vox-}\pazocal{L}_{vox-}(G) + \lambda_{c}\pazocal{L}_{c}(G, f) + \lambda_{d} \pazocal{L}_{d}(G, D) \\
    \pazocal{L}_D =  \lambda_{d-} \pazocal{L}_{d-}(G, D) + \lambda_{d+} \pazocal{L}_{d+}(D).
\end{align}
\noindent It's expected that the differences between the factual and counterfactual images reflect personalized image-based markers that are predictive of a change in the patient's disease evolution. 

\section{Experiments and Results}
To examine the effectiveness of our approach, experiments were devised to show that generated counterfactual explanations are: (1) realistic and subject specific; (2) corroborate previously established imaging markers; and (3) point toward the possibility of discovering novel imaging markers of future disease activity.

\subsection{Dataset and Implementation Details}
We leverage a large, proprietary, longitudinal RRMS clinical trial MRI dataset, with a total of 815 subjects~\cite{bravo}. There is a one-year interval between baseline and follow-up scans. For these experiments, only the Fluid Attenuated Inverse Recovery (FLAIR) sequence at the baseline time point is used as input to the network as it is the most commonly used sequence. FLAIR sequences are of size 58 $\times$ 172 $\times$ 224 voxels at a resolution of 1 $\times$ 1 $\times$ 3 mm. The dataset was divided into non-overlapping training (60 \%), validation (20 \%) and testing (20 \%) splits. Analysis and validation are performed with baseline T2 and the difference maps between the real and synthesized counterfactual image. The future MRI scans, including NE-T2 and GAD labels, were neither a part of training nor validation.

The proposed generator network architecture is a modified 4-layer nnUNet~\cite{isensee2018nnu} with a conditioned decoder. Conditions are supplied through a modulating operation $M(\cdot)$ similar to FiLM \cite{perez2018film}, performed on the features of each decoder layer.

The classifier is a 4-block ResNet~\cite{he2016deep} pre-trained on the same dataset and splits as the generator with a ROC-AUC of 0.72 on a validation set and 0.71 on a held-out test set (frozen for these experiments). The discriminator consists of a 4-block ResNet~\cite{he2016deep} which maps the generated counterfactuals
to the likelihood of the observation belonging to the real \textit{inactive} data distribution. To this end, real examples of baseline MRIs of subjects from the \textit{inactive} class with $y = 0$ are provided to $D(\cdot)$. This module is trained in conjunction with the generator and provides supervision to $G$ via $\pazocal{L}_{d}$.

\subsection{Evaluating Counterfactuals and Discovered Image-based Markers}
The absence of ground truth poses significant challenges in terms of the evaluation of the synthesized counterfactuals themselves and the corresponding image-based biomarkers that are discovered. In this work, these evaluations are based on three different criteria placed on the generated counterfactuals: That they (1) remain faithful to the subject's identity (\textit{subject-specificity}), (2) be realistic with respect to the target class in terms of adherence to clinically established, global population-level markers (\textit{target class similarity}), and (3) highlight, through difference maps with the real baseline images, local perturbations that could represent novel, personalized (as opposed to population-level) imaging markers predictive of future disease outcomes (\textit{personalized marker discovery}). Experiments will be performed to evaluate these three criteria.

For criteria (1), subject-specific faithfulness of the counterfactual image is evaluated using the average structural similarity index measure (SSIM)~\cite{hore2010image} with the input image. SSIM can take on a value between 0.0 and 1.0, with larger values indicating a higher degree of similarity.

Criteria (2) is evaluated by how well the counterfactuals reflect changes that we would expect in known image markers that have already been shown at the population-level to be correlated with future lesion activity. In the context of MS, the presence of Gad lesions and high T2 lesion volume at baseline are two attributes that are strongly correlated with future lesion activity~\cite{romeo2013clinical}. Thus, we would expect, at the population-level, to see (1) a reduction in Gad lesions and (2) a decrease in baseline T2 lesion volume in the counterfactual images, which would be in agreement with the \textit{active}-to-\textit{inactive} transition. In order to estimate these population-level markers, we use a T2 lesion segmentation network and a Gad lesion detection network. The average T2 lesion volume of the counterfactuals are estimated by a pre-trained segmentation network, that is, a nnUNet~\cite{isensee2018nnu} trained on FLAIR sequences with a DICE score of 0.60 on the hold-out test set. Second, the prevalence of Gad lesions in the counterfactual of active subjects (i.e. the generated inactive image) and in the real inactive subject images should be similar. In order to perform this validation, we pre-trained a ResNet~\cite{he2016deep} model to detect the presence of Gad lesions based on FLAIR input sequences (i.g. ROC-AUC of 0.79 on the hold-out test set).

To evaluate criteria (3), which is the model's ability to discover novel personalized markers beyond the aforementioned already established population-level markers, the difference maps between the real baseline and the counterfactual image can be analyzed qualitatively at the individual level. These difference maps highlight the perturbations, or candidate image markers, that were discovered by the deep learning model as being predictive of future disease outcomes. The candidate markers are assessed in terms of spatial location and appearance (e.g. magnitude of the perturbation). In some cases, the model may produce local changes in the baseline image that are in the vicinity of future NE-T2 or Gad lesion(s) appearing on the follow-up scan one year later, while in other cases the model may produce more global changes.

\subsection{Counterfactual Results}
\begin{figure}[h]
    \centering
    \scriptsize
    \includegraphics[width=0.8\textwidth]{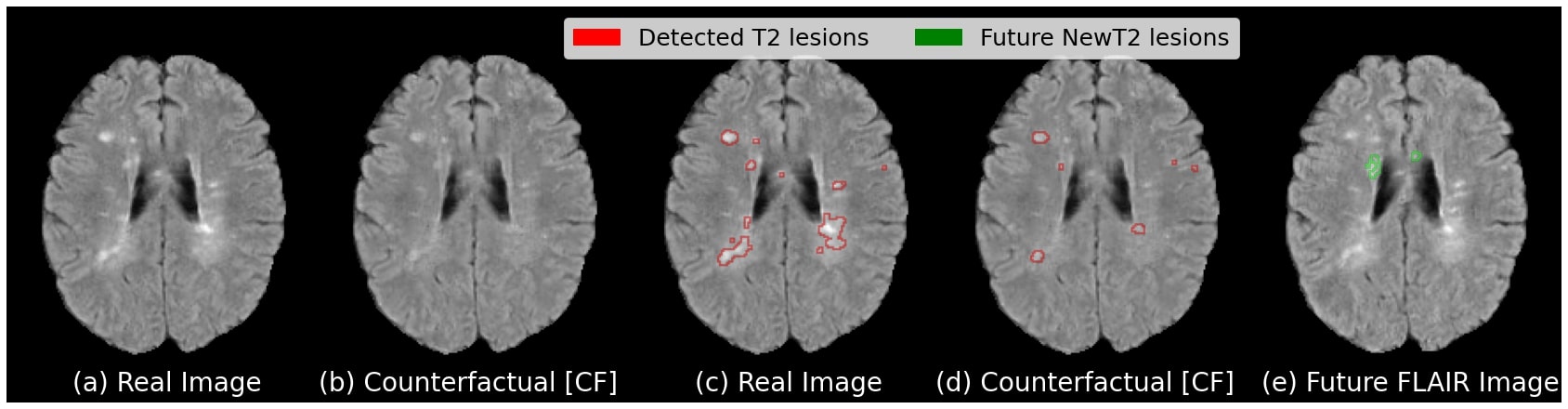}
    \caption{Example of counterfactual results for a subject. (a) Real image, (b) Counterfactual image, (c-d) Real and counterfactual image with overlayed segmented T2 lesions, (e)  FLAIR image one year later with overlayed NE-T2 lesions.}
    \label{fig:cfResult}
\end{figure}

\noindent \textbf{Subject-Specificity}
Figure~\ref{fig:cfResult} presents qualitative results for an illustrative test case. The counterfactual image in (b) shows a reduction in the number and size of the T2 lesions as compared to the real active image in (a). This can be seen by the T2 lesion segmentation maps produced by the pre-trained segmentation network, showing significantly less detected lesions in the counterfactual image (d) as compared to the real image (c). The counterfactual image maintains subject identity of the input scan with a similar overall appearance and a moderately reduced T2 lesion load. We also present the same subject's image one year later in (e), where we can see the appearance of two new T2 lesions.

In Table~\ref{tab: ssim}, we show that the omission of $\pazocal{L}_{vox-}$ results in a significant degradation in subject faithfulness of the synthesised counterfactual image. The inclusion of this loss does not impact the counterfactual validity in terms of a successful transition to the target class - all generated counterfactuals are classified as inactive by the pretrained classification network.

\newcolumntype{C}[1]{>{\centering\arraybackslash\hspace{0pt}}p{#1}}

\begin{table}[h]
\captionsetup{font=small}
\scriptsize
\centering
\caption{Effect of $\lambda_{vox-}$ on subject-faithfulness.}\label{tab: ssim}
\begin{tabular}{|c|c|c|}
\hline
\textbf{Losses} &  \textbf{Reconstruction} & \textbf{Counterfactual}\\
\hline
\begin{tabular}{C{1.2cm}|C{1.2cm}} $\lambda_{vox+}$ & $\lambda_{vox-}$ \\ \end{tabular}
& SSIM & SSIM \\

\hline
\begin{tabular}{C{1.2cm}|C{1.2cm}}  \checkmark & \checkmark \\ \end{tabular} & 0.9986 $\pm$ 0.00 &0.9667 $\pm$ 0.01\\
\hline
\begin{tabular}{C{1.2cm}|C{1.2cm}}  \checkmark &  \\ \end{tabular} & 0.9991 $\pm$ 0.00 &0.6423 $\pm$ 0.01\\
\hline
\end{tabular}
\end{table}

\noindent \textbf{Target Class Similarity}
Quantitative results depicting the average T2 lesion volumes and Gad lesion prevalence for the reconstruction and generated counterfactual image are reported in Table~\ref{tab: t2statistics}. For comparison, we also include T2 lesion volume and Gad lesion prevalence statistics for real active and real inactive samples. In the top row, we see that the average T2 lesion volume of the reconstruction is relatively close to that of the real active samples. On the other hand, the average T2 lesion volume of the counterfactual is substantially lower, being closer to the T2 lesion volume of the real inactive subjects than the active subjects. A similar trend can be observed for Gad lesion prevalence, where the Gad prevalence of the generated counterfactual has dropped closer to the prevalence observed among real inactive subjects. Overall, these results show that the counterfactual images are realistic in terms of their ability to discover clinically established population-level predictive markers, without \textit{a priori} knowledge of these markers. In this case, this leads to a reduction in baseline T2 lesions load and a removal of baseline Gad lesions from the counterfactual images.

\begin{figure}[h]

\scriptsize 
\centering
\begin{tabular}{cccc}
\centering
   \includegraphics[width=30mm]{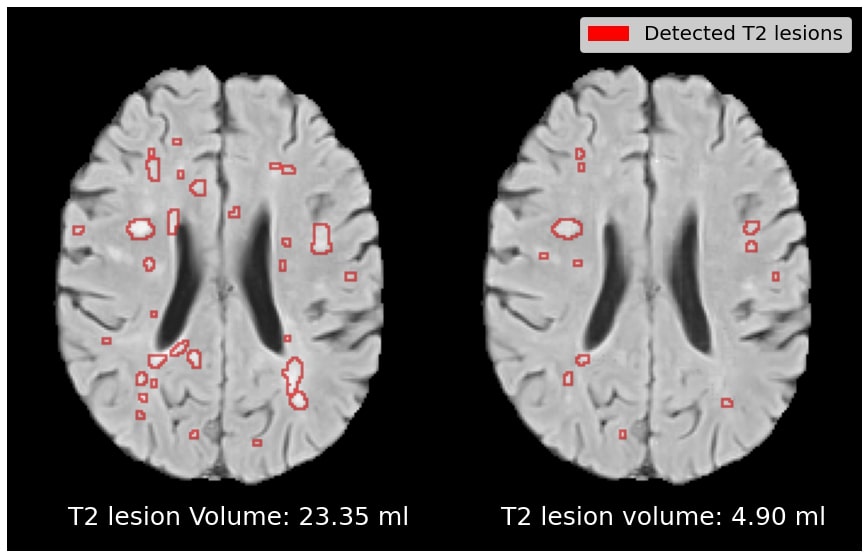} &   \includegraphics[width=30mm]{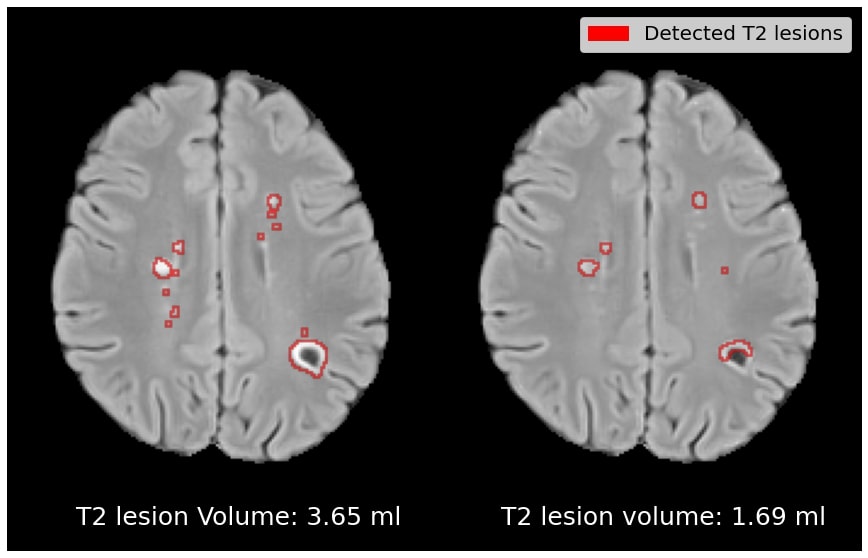} &
   \includegraphics[width=30mm]{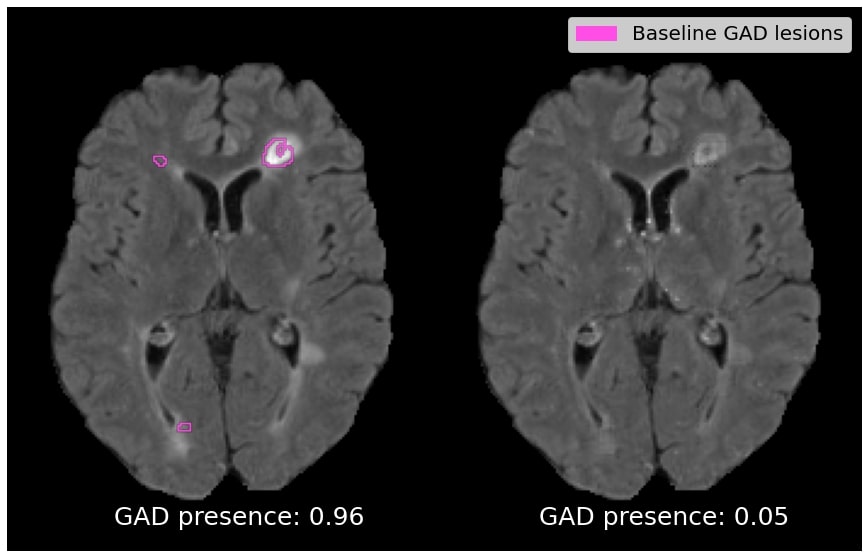} &   \includegraphics[width=30mm]{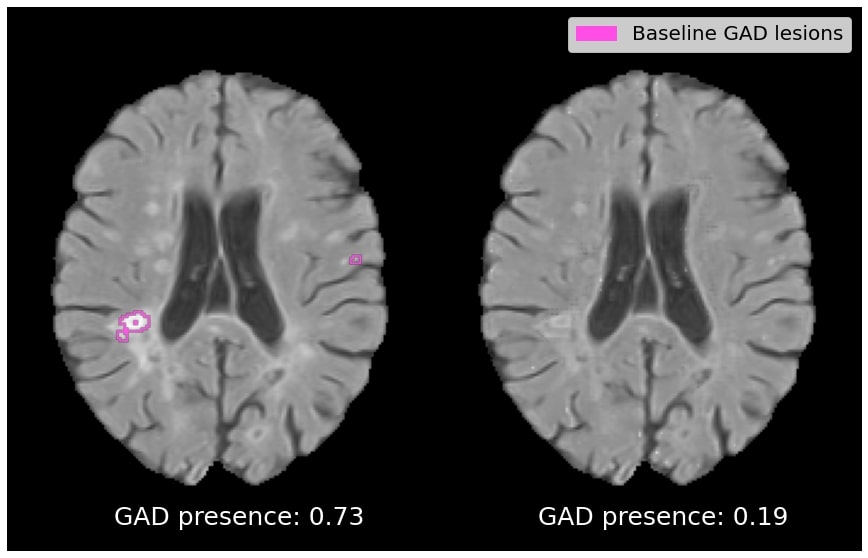}\\
  (a) & (b) & (c) & (d) \\
\end{tabular}
\scriptsize 
\caption{Validating known image markers observed in counterfactual images (right image in (a)-(d)) against input images (left image in (a)-(d)). (a)-(b): reduction of lesion load in the counterfactual image, the change in the T2 lesion volume is shown under each image; (c)-(d): removal of baseline Gad lesions in the counterfactual and results from the pre-trained gad presence detector are reported at the bottom of each image.}
\label{fig:gadBaseline_lesionLoad}
\end{figure}

Figure~\ref{fig:gadBaseline_lesionLoad} depicts qualitative results for the effect on T2 and Gad lesion load. In (a), we show an example of a patient with above-average lesion load, with the counterfactual exhibiting a 79\% drop in T2 lesion load. In (b), an active patient is shown with a below-average lesion load (i.e. an example that goes against the population-level trend), with the counterfactual exhibiting a 54\% drop in T2 lesion load. In (c) and (d), the counterfactual images exhibit a complete suppression of Gad lesions. Overall, Gad lesions are suppressed to a much greater extent than T2 lesion suppression. This aligns with the statistics for real \textit{active} and real \textit{inactive} populations (see Table~\ref{tab: t2statistics}).

\begin{table}[h]
\scriptsize
\centering
\caption{Baseline lesion statistics for real and generated samples}\label{tab: t2statistics} 
\begin{tabular}{|c|c|}
\hline
 & \begin{tabular}{C{2.2cm}|C{2.2cm}|C{2.2cm}|C{2.2cm}} \textbf{Real (Active)} $\mathbf{x} \sim \pazocal{X}_{y=1}$&\textbf{Reconstruction} $G(\mathbf{x}, 1)$ &\textbf{Counterfactual} $G(\mathbf{x}, 0)$ &\textbf{Real (Inactive) $\mathbf{x} \sim \pazocal{X}_{y=0}$}\end{tabular}\\
\hline
\textbf{Avg. T2 lesion vol.}&\begin{tabular}{C{2.2cm}|C{2.2cm}|C{2.2cm}|C{2.2cm}}
$13.40\pm13.76$&$14.37\pm14.32$& $2.27\pm3.91$  &$7.20\pm 8.23$ \end{tabular}\\ 
\hline
\textbf{Gad Prevalence}&\begin{tabular}{C{2.2cm}|C{2.2cm}|C{2.2cm}|C{2.2cm}}
$33.87\%$ & $31.37\%$ & $0\%$  & $3.43\%$ \end{tabular}\\ 
\hline
\end{tabular}
\end{table}

\begin{figure}[h]
\centering
\scriptsize
\begin{subfigure}[b]{0.49\textwidth}
\centering
\includegraphics[width=\textwidth]{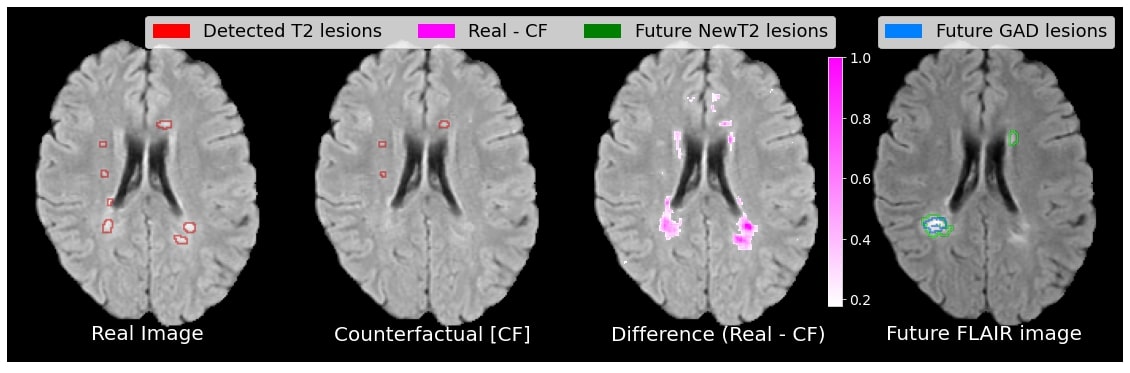}
\caption{}
\label{fig:l1}
\end{subfigure}
\hfill
\begin{subfigure}[b]{0.49\textwidth}
\centering
\includegraphics[width=\textwidth]{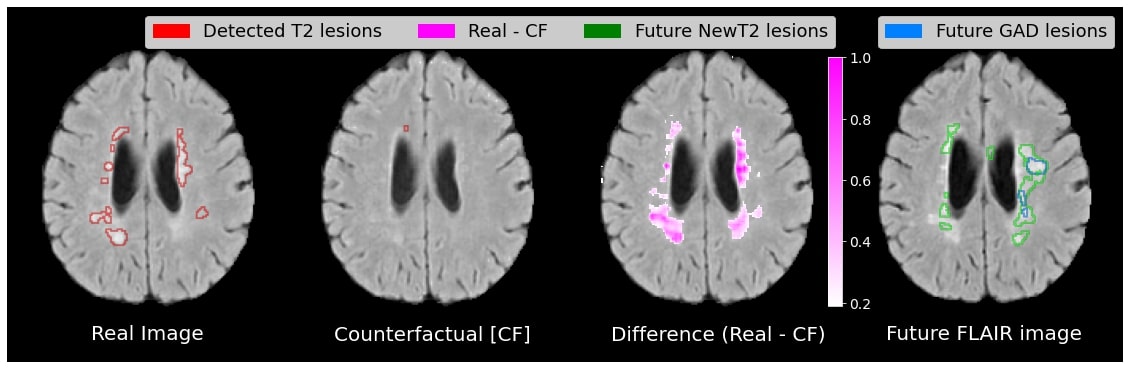}
\caption{}
\label{fig:l2}
\end{subfigure}
\scriptsize
\caption{Subject-specific image-marker discovery using difference maps. FLAIR images are shown for two different subjects (a and b). From Left to right: real image, counterfactual image, real image with the intensity of the difference map (real minus CF) overlayed in magenta,  and future flair image (one year later) with overlayed NE-T2 (green) and Gad lesions (light blue). Note that, in some cases, the difference maps indicate areas where future NE-T2 or Gad lesions appear.}
\label{fig:subject-specific}
\end{figure}

\noindent \textbf{Personalized Marker Discovery}
Using a personalized approach to predictive image marker discovery, we would expect different individuals to have predictive markers in different locations, and potentially of a different type or appearance (e.g. in terms of magnitude of the perturbation). We would also expect these personalized markers to point to potentially novel processes beyond the aforementioned established population-level markers (T2 lesion volume and Gad count, in the case of MS). 
A qualitative analysis of different test cases demonstrating subject-level heterogeneity in the candidate image markers is shown in Figure~\ref{fig:subject-specific}.
The candidate markers for subject (a) are found in a variety of locations including periventricular, deep white matter, and subcortical/juxtacortical, and cortical regions, while the candidate markers for subject (b) are localized to the periventricular region. In both cases, some perturbations occur in the region where future new T2 lesions will appear (or where existing T2 lesions will enlarge), indicating that the model has identified local predictive markers of future lesion activity. These changes are either in regions where there are T2 lesions in the real baseline image, or in regions where there are no lesions. This is in line with current understanding of how some lesions (enlarging lesions) enlarge slowly over time~\cite{Elliott2019}, and how non-lesional tissue called ``diffusely abnormal white matter'' can later transform into new/de novo T2 lesions~\cite{Dadar2022}. Moreover, both cases also show candidate markers in brain regions distant to the area where NE-T2 will occur, indicating the discovery of global (rather than local) predictive markers of future activity. This is particularly noticeable in the cortex of subject (a), just anterior to the lateral ventricles. These global markers could represent a general predisposition to having more NE-T2 lesion activity (just as T2 lesion load in general is predictive of future lesion activity), but could also imply the importance of certain specific brain regions, tracts, or networks, in MS lesion development. These candidate markers are meant to form the basis of future scientific inquiry and can be the subject of dedicated validation studies.

\section{Conclusions}

Explainable deep learning models that enable data-driven discovery of patient-specific imaging markers predictive of future patient outcomes would lead to a better understanding of individual disease evolution, and would also increase the likelihood of adoption of deep learning models in the clinic by increasing their trustworthiness. In this work, we propose a new deep generative model for counterfactual image synthesis to aid in the discovery of personalized biomarkers predictive of future disease outcomes. By perturbing factual baseline images, the model allows for the visualization of image markers that explain the switch to an alternate future disease state. The quantitative evaluations show that the synthesized counterfactuals maintain subject specificity and target class similarity. A qualitative analysis of candidate markers reveals heterogeneity in the spatial location and appearance of candidate markers of different individuals, underscoring the value of using a personalized approach to image marker discovery. The discovered patient specific markers were challenging to validate quantitatively, given that there is no established ground truth. Future work is required to further analyse the difference maps in order to uncover trends among patient sub-populations. Extending our approach to patients with closer MRI scans (less than one year), would also be interesting, as it would potentially enable the discovery of additional local markers in the vicinity of future NE-T2 lesions.

\subsubsection{Acknowledgements} The authors are grateful to the International Progressive MS Alliance for supporting this work (grant number: PA-1412-02420), and to the companies who generously provided the clinical trial data that made it possible: Biogen, BioMS, MedDay, Novartis, Roche / Genentech, and Teva. Funding was also provided by the Natural Sciences and Engineering Research Council of Canada, the Canadian Institute for Advanced Research (CIFAR) Artificial Intelligence Chairs program, and a technology transfer grant from Mila - Quebec AI Institute. S.A.\ Tsaftaris acknowledges the support of Canon Medical and the Royal Academy of Engineering and the Research Chairs and Senior Research Fellowships scheme (grant RCSRF1819 /\ 8 /\ 25). Supplementary computational resources and technical support were provided by Calcul Québec, WestGrid, and Compute Canada. This work was made possible by the end-to-end deep learning experimental pipeline developed in collaboration with our colleagues Justin Szeto, Eric Zimmerman, and Kirill Vasilevski. Additionally, the authors would like to thank Louis Collins and Mahsa Dadar for preprocessing the MRI data, Zografos Caramanos, Alfredo Morales Pinzon, Charles Guttmann and István Mórocz for collating the clinical data, Sridar Narayanan, Maria-Pia Sormani for their MS expertise.
%
%
%
\newpage
\bibliographystyle{splncs04}
\bibliography{paper17}

\end{document}